\documentclass[runningheads]{llncs}
\usepackage{graphicx}
\usepackage{tikz}
\usepackage{comment}
\usepackage{amsmath,amssymb} % define this before the line numbering.
\usepackage{color}
\usepackage{amssymb}
\usepackage{booktabs}
\usepackage{enumitem}
\usepackage{verbatim}
\usepackage{float}
\usepackage{hyperref}
\usepackage[caption=false]{subfig}

\newlength\savewidth

\usepackage{pifont}% http://ctan.org/pkg/pifont

\usepackage{xcolor}
\usepackage{colortbl}

\begin{document}
% \linenumbers
\pagestyle{headings}
\mainmatter
\def\ECCVSubNumber{2670}  % Insert your submission number here

\title{Answer-Me: Multi-Task Learning for  Generalization to Many Question-Answering Tasks }
\title{Multi-Task Learning for  Generalization to Many Question-Answering Tasks }
\title{Multi-Task Generalization for Visual Question Answering}

\title{Answer-Me: Multi-Task Open-Vocabulary Visual Question Answering}

%\begin{comment}
\titlerunning{Answer-Me} 
\authorrunning{AJ Piergiovanni et al.} 
\author{AJ Piergiovanni, Wei Li, Weicheng Kuo,  Mohammad Saffar, Fred Bertsch and Anelia Angelova}
\institute{Google Research}
%\end{comment}
%******************

\begin{comment}
\titlerunning{Abbreviated paper title}
\author{First Author\inst{1}\orcidID{0000-1111-2222-3333} \and
Second Author\inst{2,3}\orcidID{1111-2222-3333-4444} \and
Third Author\inst{3}\orcidID{2222--3333-4444-5555}}
%
\authorrunning{F. Author et al.}
\institute{Princeton University, Princeton NJ 08544, USA \and
Springer Heidelberg, Tiergartenstr. 17, 69121 Heidelberg, Germany
\email{lncs@springer.com}\\
\url{http://www.springer.com/gp/computer-science/lncs} \and
ABC Institute, Rupert-Karls-University Heidelberg, Heidelberg, Germany\\
\email{\{abc,lncs\}@uni-heidelberg.de}}
\end{comment}
%******************
\maketitle

\begin{abstract}
We present Answer-Me, a task-aware multi-task framework which unifies a variety of question answering tasks, such as, visual question answering, visual entailment, visual reasoning. In contrast to previous works using contrastive or generative captioning training, we propose a novel and simple recipe to pre-train a vision-language joint model, which is multi-task as well. The pre-training uses only noisy image captioning data, and is formulated to use the entire architecture end-to-end with both a strong language encoder and decoder. Our results show state-of-the-art performance, zero-shot generalization, robustness to forgetting, and competitive single-task results across a variety of question answering tasks. Our multi-task mixture training learns from tasks of various question intents and thus generalizes better, including on zero-shot vision-language tasks. We conduct experiments in the challenging multi-task and open-vocabulary settings and across a variety of datasets and tasks, such as VQA2.0, SNLI-VE, NLVR2, GQA. We observe that the proposed approach is able to generalize to unseen tasks and that more diverse mixtures lead to higher accuracy in both known and novel tasks. 

%\keywords{Visual Question Answering, Image-Language Learning}
\end{abstract}

%We will open source the code.

%Answer-Me presents a model and pretraining method for learning image and natural language interaction for a variety of tasks, such as Describe the image (Captioning), Visual Question Answering, visual entailment. Specifically, we show that for vision-language tasks, an Encoder-Decoder model is better than Decoder-only style models and that by casting the tasks as a generative text setting, the model is able to generalize to many tasks, including zero-shot vision-language tasks. We also present a pretraining method using only noisy image captioning data but formulate the training so that it is able to train both a strong language encoder and decoder. We conduct experiments acorss a variety of vision-language datasets and tasks such as VQA, SNLI-VE, and CocoCaptions.

%We demonstrate multi-modal multi-task learning for natural language interaction with images with responses to questions from a variety of tasks, such as Describe the image (Captioning), Visual Question Answering, visual entailment. The interaction should feel natural and meaningful. We propose task-aware multi-task pretraining and training. We show that our models understand the query intent automatically and generalize to unseen tasks. Stress on image-language feature interaction.

%\end{abstract}

%%%%%%%%% BODY TEXT
\section{Introduction}

The ability to understand both visual and textual cues is crucial for interactions grounded in the rich visual world. 
Questions are a natural way for users to articulate information needs,
and a variety of tasks in which a question is posed towards an image, such as Visual Question Answering (VQA), % image captioning, %image-to-text and text-to-image retrieval, 
visual commonsense reasoning, visual entailment and others, have been created~\cite{agrawal2015vqa,goyal2017making,hudson2019gqa,snli,snli-ve,zellers2019vcr,gd-vcr}.
Question answering with an image is a challenging task as it involves deeper visual understanding, for example `What is the child holding?' requires visual recognition, whereas `Why is the person smiling?' requires additional reasoning, beyond just recognition.
%In this work, we explore a generalized VQA task by requiring a single model to be able to answer all types of questions.
%In this work, we consider the generalized task of natural language interaction with images combining the above-mentioned tasks.
We build a multi-task model which can use natural language to ask questions about an image and respond in free-form text, enabling the model to answer many questions, even out-of-domain ones.

\begin{figure}%[t]
    \centering
    \includegraphics[width=1.0\linewidth]{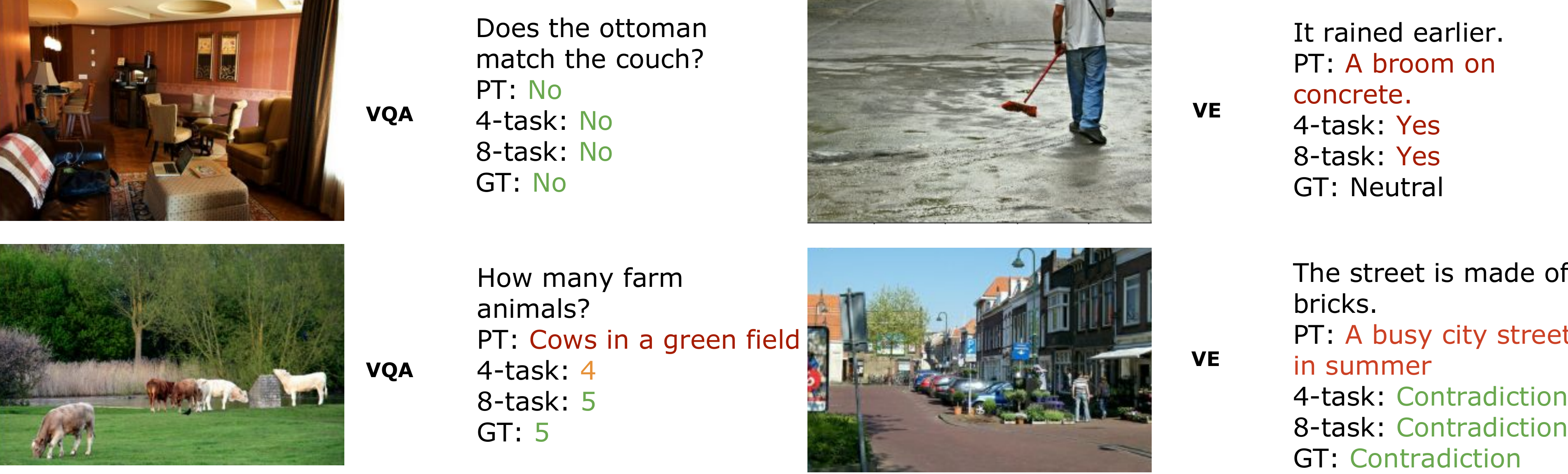}
    \caption{Examples comparing zero-shot performances between a pretrained (PT) model on a large dataset, our multi-task learning on 4 tasks, and our multi-task learning on 8 tasks, evaluated on VQA and visual entailment (VE) tasks. All are zero-shot performances with respect to the new question and answering task: while pretrained models are powerful they lack understanding of the question intent and are not able to respond to questions as adequately as our task aware multi-task setup.}
    \label{fig:motivation}
    \vspace{-0.5cm}
\end{figure}

%\textbf{Some motivation:}
%Previous multi-modal image-language approaches face several challenges.
%With a plethora of 
Multi-modal image-language approaches have made great strides recently, showing tremendous promise in joint visuo-linguistic understanding. 
Most commonly, they use pretraining on large generic datasets, and then fine-tuning on multiple downstream tasks~\cite{vilbert2020,ernievil,chen2020uniter,align,su2020vlbert,li2020oscar}. Pretraining is appealing because it can be done once and adapted to multiple tasks.
However pretraining is {\it task agnostic}, i.e., is not tuned to the tasks' goals, and finetuning is often done for each individual task independently, creating multiple, different copies of the model. 
Importantly, the intent of the input text is not captured by many pretrained models, as they are trained on datasets for image captioning %or description 
or 
with weak image-language correlations~\cite{clip,align}, which might not learn the right interactions between image and language to capture the goal of the question. For example, pretraining with image captioning does not train a text encoder, so the model will not be able to understand questions. Instead, we need a model which is able to take advantage of the language-image training and learn the corresponding interactions which reflect the intent of the question or task.

Furthermore, it is desirable that image-language models are able to generalize to other tasks with natural questions and answers, which are not seen during training. This is needed because models fine-tuned on a specific task tend to demonstrate larger rates of catastrophic forgetting on new tasks. For example, GPV~\cite{gupta2021gpv} demonstrated that fine-tuning on one task leads to a loss of 50\% precision on another task, after adding only about a hundred examples. This overfitting and forgetting is observed even if very large datasets are used for pretraining; we also observe it here (Section~\ref{sec:across}).

Lastly, architecture complexity has been prevalent, with complex box proposal mechanisms used to obtain the initial image features making the architectures cumbersome to train and support. This limits the generalization abilities of the model in cases where objects are not detected (e.g., out-of-domain objects) and prevents models from general adoption, dissemination and makes them harder to scale.

We propose `Answer-Me' which unifies visual question answering tasks and aims to answer a variety of  natural language questions towards an image from a wide range of tasks. The gist of the method is multi-task, task-aware training, which is able to take in the question intent. This is combined with our novel pretraining which learns the encoder, decoder and image-language feature interaction end-to-end, and is also multi-task itself. The pretraining method is simple; it uses only image captioning data in multiple ways in order to train both the image and language encoders and the text decoder model components. This allows for  natural language questions and free-form answers that recognize intent and answer accordingly, without additional prompts.
We evaluate our multi-task approach in the open-vocabulary setting which is more challenging, as the generated text has to match exactly the ground truth to be counted as accurate, as opposed to other approaches which use classification as a proxy, choosing among a pre-defined set of possible answers. We note that open-vocabulary text generation is compounded by the multi-task setting, as the method does not have the opportunity to fine-tune on specific tasks, thus it cannot easily memorize specific answers within a dataset. While challenging, this is a more realistic scenario and thus more valuable to sustain in evaluations.

%We demonstrate understand the intent automatically and generalize to unseen tasks.
The approach is simple and easily extensible to more tasks and datasets and is shown to generalize to novel tasks. %and scalable.
It performs well compared to the state-of-the-art (SOTA), largely outperforming prior multi-task methods and in some cases pre-trained and fine-tuned ones, despite the more challenging open-vocabulary evaluation setting.

We find the following advantages of the proposed multi-task training:
   (i) semantically related tasks benefit from multi-task training as visual linguistic concepts are shared, helping their co-training;
   (ii) pooling tasks increases diversity of the questions, allows the model to react to different question types, leading to improved generalization to novels tasks;
   (iii) training multiple tasks improves performance, 
   reducing catastrophic forgetting and overfitting.

%This work does not propose a new model, instead we simplify the architecture as much as possible to address a union of QA tasks. 
We make the following contributions:
\begin{itemize} %[leftmargin=*]
\setlength\itemsep{0.00em}
\item A multi-task task-aware general framework which can serve multiple diverse Visual Question Answering tasks and is able to %understand the intent and
generalize to many tasks. It uses a simple, easily scalable and extensible architecture and considers the VQA tasks in the more-challenging open-vocabulary generative setting. 
\item A pretraining method to train the entire encoder-decoder vision-language model simultaneously using only noisy captioning data that results in a strong pretrained model.
\item Answer-Me is able to perform well on zero-shot (novel task) generalization settings, is also robust to overfiting and forgetting, and performs well on a new `detection' task which requires semantic object recognition.
\end{itemize}

%%%%%%%%%%%%%%%%%%%%%%%%%%%%%%%%%%%%%%%%%%%%%%%%%%%%%%%%%%%%%%%%%%
\vspace{-0.2cm}
\section{Related work}
\vspace{-0.05cm}
\textbf{Multi-modal image-language learning.}
Multi-modal image-language learning has garnered large interest in recent years, encompassing tasks, such as VQA,~\cite{agrawal2015vqa,goyal2017making,gurari2018vizwiz,jiang2020in,whitehead2021separating,hudson2019gqa}, visual commonsense reasoning ~\cite{suhr2017acorpus,snli-ve,zellers2019vcr}, visual dialog~\cite{kottur2018Visual,das2017Visual} image captioning~\cite{chen2015cococaptions,anderson2018bottomup,cc3m,beer2021cc12m}, cross-modality retrieval i.e., image-to-text and text-to-image~\cite{clip,align}, referring expressions comprehension~\cite{mao2016generation,yu2016modeling,yu2018mattnet,qiao2020referring,yang2019fast,mdetr,liu2019learning,liao2020real,deng2021transvg} and many others~\cite{plummer2017FlickrEntities,Rohrbach2016grounding,plummer2020revisiting,yang2020improving,Tuay1018dynamic}. 
Some tasks use joint image and language inputs, such as captioning or VQA~\cite{cc3m,beer2021cc12m,chen2015cococaptions,agrawal2015vqa,goyal2017making,gurari2018vizwiz}, while
some use image and language learning to joint space which enables cross-modal retrieval~\cite{align,clip,srinivasan2021wit}.
%Clip also leverages language to extend the class labels for purely visual recognition task~\cite{clip}. 
Additional modalities are also often incorporated in such models, for example, video, audio, or additional text such as transcriptions or video captions~\cite{xu2016mstvtt,rohrbach2015adataset,deng2021sketch,huang2020multimodal,lin2021vx2text}.

\textbf{Transfer learning for image-language models.}
A common approach to leveraging image and text datasets for multiple tasks is to pretrain on a related dataset, %(but not necessarily solving a specific task) 
and then fine-tune on downstream datasets and tasks. Pretraining from large data has shown impressive performance gains: %on downstream tasks:
%Many works have leveraged pretraining for improved downstream tasks:
e.g., ViLBERT~\cite{vilbert2020} pretrain on the Conceptual Captions dataset~\cite{cc3m,beer2021cc12m}, % in a class-agnostic manner %
and with some modifications to the architecture, fine-tune on a number of tasks.
Other works also show successful training on downstream image-and-language tasks: VLBERT~\cite{su2020vlbert}, LXMERT~\cite{tan2019lxmert}, UNITER~\cite{chen2020uniter}, Oscar~\cite{li2020oscar}, SOHO~\cite{huang2021seeing}, Unified VLP~\cite{zhou2020unifiedVLP}, VisualBERT, ~\cite{beer2021cc12m}, 
%(pretrain on CC12m on two: captioning and matching, downstream retrieval tasks e.g. nocaps (for captioning), flickr (for matching) - note that different pre-tr matched different site), 
%~\cite{sariyildiz2020learning}, % (COCO & VG), 
ViLT~\cite{kim2021vilt}, % (pretr- coco, cc, vg, mask modeling objectives, uses VT) eval on VQA and NLVR2, and COCO and Flickr for retrieval.
and others~\cite{align,srinivasan2021wit,huang2021seeing,villa,ernievil,sariyildiz2020learning} %pretrain on MSCoco and VG Downstream Datasets/Tasks: VQA, Content Reasoning (NLVR), Ima-Text Retrieval (TR/IR, COCO and Flickr), Visual Entailment (VE)
%In some of these works~\cite{align,srinivasan2021wit,beer2021cc12m}, pretraining can be even successfully done from large but noisy datasets and the results on downstream or zero-shot tasks are not affected. 

Pretraining techniques, as such, also vary. In some cases pretraining is done by training on a single task e.g., captioning and a large dataset~\cite{cc3m,vilbert2020}, others use different pretraining objectives depending on the downstream task, for example,~\cite{beer2021cc12m} uses captioning and matching losses for pretraining, where the former is used for downstream VQA or captioning tasks, % e.g. nocaps~\cite{nocaps}, 
and the latter for retrieval tasks. %flickr (for matching) - note that different pre-tr matched different site),  
Contrastive pretraining is particularly useful for retrieval tasks~\cite{clip,align}.
Masked modeling objectives \cite{devlin2018bert}, and its derivatives, is one of the most popular form of pretraining for joint image and language learning. Following the same idea for text~\cite{vilbert2020,sun2019videobert,su2020vlbert,chen2020uniter,cho2021unifying}, %, multiple ideas have been proposed for masking and matching image regions (or pre-detected object regions), and text words or tokens~\cite{kim2021vilt}.
%Other works, such as
VirTex~\cite{virtex} and LocTex~\cite{loctex} also show that pretraining on both image and language modalities jointly benefits vision-only tasks. %In our work, we use pretraining, where our pretraining is fully leveraged by the multi-task question answering tasks. 

\textbf{Image language learning.}
Many image-language interaction learning methods~\cite{vilbert2020,mdetr} are based on the the popular Transformer architecture~\cite{vaswani2017attention}.
For example, co-attention with Transformer in VilBERT~\cite{vilbert2020}.
Nguyen et al.~\cite{nguyen2019multi} proposed attention-based image-language %interaction 
approach and~\cite{changpinyo2021telling} use it for localization.
%Changpinyo et al. 

%shows that better retrieval can be accomplished when additional localization information in this case from mouse traces is provided. % during training. 
%Our approach builds on these works, and proposes also image-language interaction modeling.

\textbf{Multi-task learning for vision and language.} 
Our work is most aligned with the image-language multi-task 
approaches~\cite{nguyen2019multi,liu2019multitask,pramanik2019omninet,lu202112in1,gupta2021gpv,cho2021unifying,hu2021unit,shuster2020dodecathlon}. In early work, Nguyen et al.~\cite{nguyen2019multi} combine three vision and language tasks by incrementally adding tasks to the training. %, reusing shared components. %image-language interaction.
Following the success of unified multi-task learning for text, e.g., T5~\cite{T5}, MT-DNN~\cite{liu2019multitask}, a number of vision+language multi-task approaches have been proposed~\cite{lu202112in1,gupta2021gpv,cho2021unifying,hu2021unit}
Cho et al.~\cite{cho2021unifying} unify several vision and language tasks as a text-based generation task. GPV~\cite{gupta2021gpv} combines tasks with diverging outputs, e.g., VQA, captioning, localization. UniT~\cite{hu2021unit} proposes multi-task learning which spans many tasks. 12-in-1~\cite{lu202112in1} train jointly several diverse vision and language tasks. Many of the multi-task approaches mentioned above rely on off-the-shelf object bounding boxes during training, or detection-specific pretraining ~\cite{lu202112in1,hu2021unit,vilbert2020,beer2019decoupled,cho2021unifying}, which might create burdensome architectures and additional complexities in data gathering and training. 
12-in-1~\cite{lu202112in1} %, based on Vilbert~\cite{vilbert2020},
utilizes FasterRCNN~\cite{faster} for detection,~\cite{gupta2021gpv,cho2021unifying} utilize DETR~\cite{detr}, whereas UniT~\cite{hu2021unit} uses a ResNet architecture, pretrained on detection. Further \cite{cho2021unifying} relies on ground truth boxes for many of the training tasks. Our work uses simple architectures without bounding boxes and relies instead on learnable image-language interactions. %Further, approaches, such as 12-in-1 and UniT use multiple, different task `heads' to learn the specific outputs for each task, whereas Answer-Me uses a single, generic text output layer, capable of generating any answer.

In complement to the above-mentioned multi-task learning approaches,
our main motivation for this work is creating a general visual representation which can be directly utilized by multiple diverse image-language tasks. Our model is able to respond naturally depending on the intent and the implied task from the question.  
It is easily extensible to many tasks in the mix and new unseen tasks, it is also easily scalable as it does not rely on complex mechanisms. 
%adapted to multiple diverse downstream tasks. 
%Instead we create both representation and approach which can solve a variety of language vision interaction tasks.
%e.g.~\cite{cho2021unifying}.

%%%%%%%%%%%%%%%%%%%%%%%%%%%%%%%%%%%%%%%%%%%%%%%%%%%%%%%%%%%%%%%%%%%%%%%%%%%%%
\vspace{-0.15cm}
\section{Multi-task Learning for Visual Question Answering}
\label{sec:main}
\vspace{-0.05cm}
\subsection{Tasks and query intent}
Each of the image-language question-answering tasks contain a specific intent in the question, for example, counting the number of objects, answering a visual entailment question, or reasoning about an image.
% Do we need examples e.g. some might be answering a counting question  `How many ducks are there?', `Are the ducks twice as many as the chickens?',
%some answering a recognition or a visual entailment question, 
Answer-Me combines the tasks in joint training using a mixture, enabling the model to acquire new capabilities, and generalize to other classes of questions without losing accuracy on tasks.

\begin{figure}[t]
    \centering
    \includegraphics[width=0.7\linewidth]{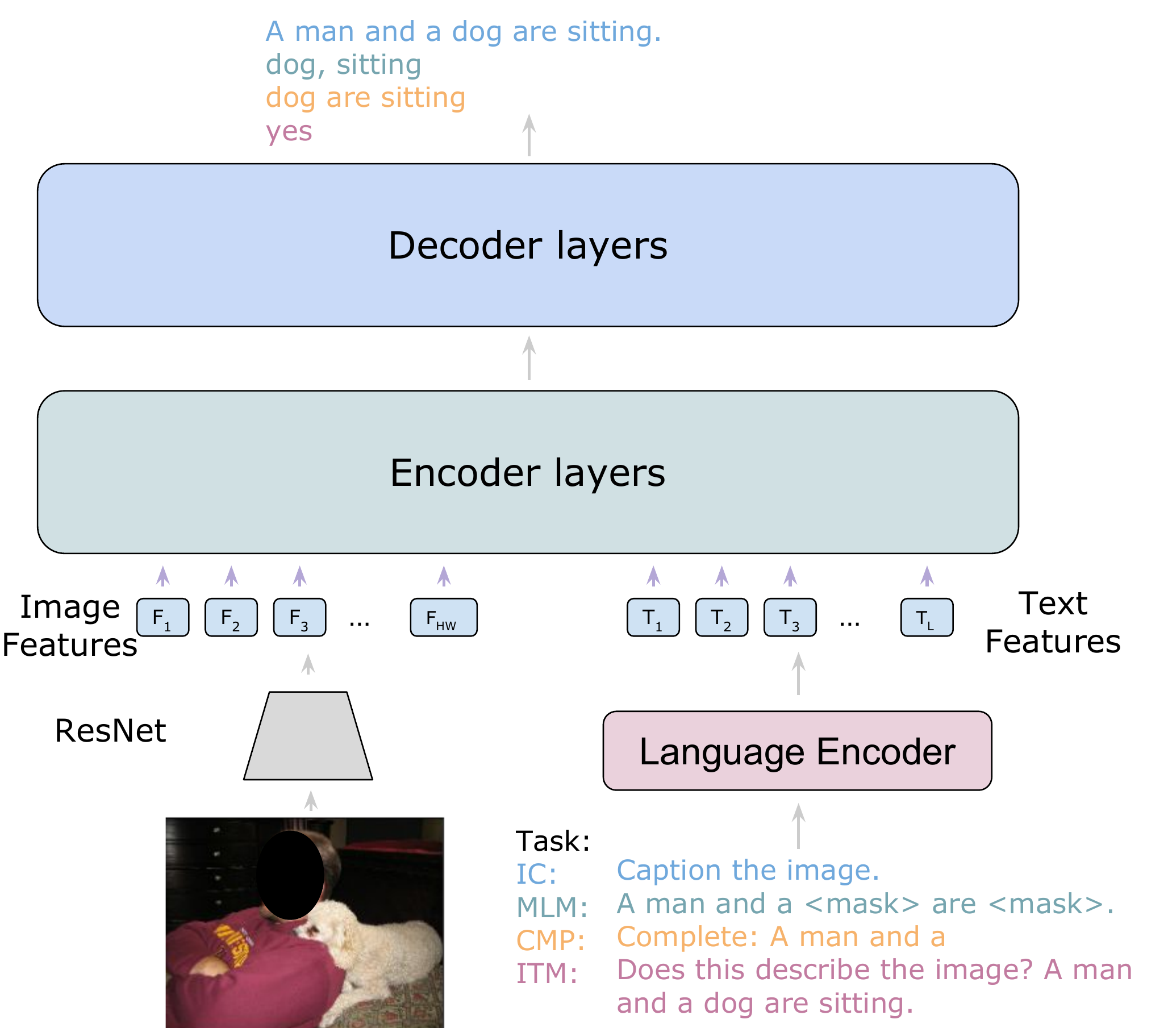}
    \caption{Model overview. The model processes an image and text, fuses them together, and uses transformer layers to generate the output text. The pretraining task mix, shown in different colors at the bottom right (see Section~\ref{sec:pretr}) allows all parts of the model to be well-trained, and is better suited for the subsequent multi-task training.}
    
    \label{fig:arch}
    \vspace{-0.4cm}
\end{figure}

In contrast to prior work which leverages large data, task-agnostic pretraining, but uses only partially pretrained models, Answer-Me both 1) pretrains specifically to exercise all components of the model with a related image-language task, and 2) trains with multiple datasets with various question intents, specifically to increase generalization. 
For example, even with very large pretraining, when finetuning on a single task, e.g., captioning or VQA, the model tends to overfit to the task itself and does not generalize to other tasks or question types. %GPV~\cite{gupta2021gpv} showed catastrophic forgetting even only after a hundred new examples with a pretrained model. 
By mixing tasks together, the model is able to better generalize to other datasets and tasks, even in a zero-shot setting.
Section~\ref{sec:arch} describes the overall architecture, Section~\ref{sec:pretr} describes the pretraining strategy which trains the model end-to-end to exercise different components.
Section~\ref{sec:multi-task} details the multi-task training.

\subsection{Main architecture}
\label{sec:arch}
Our model consists of an image encoder (ResNet) and text encoder (T5~\cite{T5}). These components are easily and independently scalable. Our experiments are based on a ResNet-50 and T5-base model, and we scale it 3x by using ResNet-101 and T5-large. %and the 3x model uses ResNet-101 and T5-large. 
The image and language features are provided to the fusion module, described below. The output of the fusion module is used as input to the text decoder, which produces free-form text for all Answer-Me tasks. The architecture is designed to be modular and easily scalable, as shown in our experiments. %We scaled our model from 300M to 1B parameters, which improved performance.
\autoref{fig:arch} shows an overview.

\paragraph{Image-language fusion.}
%\label{sec:fusion}
%
The image and language features are first fused by concatenation. 
We use the ResNet feature map to get $H*W$ features and concatenate with the $L$ (number of tokens) text features. We apply a relative position bias to both image and text separately.  We then apply the Transformer~\cite{vaswani2017attention} on the concatenated features, combining both sources. While existing works have proposed similar fusion methods,
%this fusion is not novel 
(e.g., \cite{chen2020uniter,cho2021unifying}), the pretraining method and generalisation abilities without forgetting are new, further only using raw images instead of region proposals.

\subsection{Pretraining for multi-task learning}
\label{sec:pretr}
In order to enable the model to address new tasks, i.e., to respond to unseen question types and answer adequately, we take advantage of a unique pretraining designed to train all the components of a model. We pretrain on the Conceptual Captions 12 million (CC12m) dataset~\cite{beer2021cc12m}. Unlike previous work, this pretraining strategy is targeted towards training the entire encoder-decoder model, and differs by the following components:

\vspace{-2mm}
\begin{itemize} [leftmargin=*]
\item It is designed to train all components of the model end-to-end. Thus the image encoder, text encoder, image-text fusion module and text decoder are all trained together.
\item Pretraining exercises various pathways in the model, which makes it suitable for various question-answering and description tasks. We construct a mix of captioning, caption completion and matching tasks for that purpose.
\item Taken together, these tasks allow the encoder and decoder to see all, part of, or none of the caption, and experimentally we find this pretraining method results in a stronger model than any individual pretraining task (see \autoref{tab:pretraining-tasks}).
\end{itemize}

We pretrain the model as follows. For each sample, we have an (image, text) pair, obtained from the CC12m dataset. To train all parts of the model, we design four tasks (shown in \autoref{fig:arch}):
\textbf{(1) image captioning (IC).} Here the input text is `caption the image' and the target text is the caption. This task mostly trains the text decoder and fusion layers.
\textbf{(2) caption completion (CMP).} Here the input is 10-40\% of the caption text and the target text is the remaining caption. This encourages training of the entire model.
\textbf{(3) text MLM \cite{devlin2018bert}.} Here the input is the caption with 25\% of the words masked out, the target text is the missing words\footnote{We also tried generating the entire caption and it performed similarly.}. This trains the entire model.
\textbf{(4) image text matching (ITM) \cite{chen2020uniter}.} Here the input is either the image caption or a random caption and the target text is `true' or `false' depending on if the caption matches the image or not. This primarily trains the encoder and fusion layers.

When training together, these tasks allow the model components, e.g., encoder, decoder, etc., to see all, part of, or none of the caption, and thus prepare the final model to address a variety of image-text tasks.

Another key advantage of this approach is that the training loss is simple: cross entropy over the tokens. All tasks use the same loss, including target tasks such as VQA, visual entailment, etc. %Only the preprocessing changes.

\vspace{-0.15cm}
\subsection{Multi-task training}
\label{sec:multi-task}
The multi-task training is done by taking a set of $N$ tasks and mixing them together, sampled so that a batch consists of an equal amount of each dataset, i.e., batch size/$N$ samples from each task. Since we use a text generation setting for the tasks, the loss is computed over the tokens, all using the same vocabulary. We use the same vocabulary as T5 \cite{T5} with 32,000 tokens for all our experiments. %The supp. materials have more details.

\vspace{-0.15cm}
\section{Experimental setup}
\vspace{-0.05cm}
\subsection{Datasets and tasks}
We use the following datasets to address a number of question-answering tasks:

\textbf{Visual Question Answering (VQA2.0).} We use the popular Visual Question Answering dataset VQA2.0 dataset~\cite{agrawal2015vqa}, which has about 400k training question/answer pairs. Here we use the words as the output tokens.

\textbf{Visual Entailment (SNLI-VE).} SNLI-VE ~\cite{snli-ve} is a Visual Entailement (VE) dataset with $\sim$500k samples. The VE task involves reasoning about the image and concluding whether a proposed statement is `entailment', a `contradiction', or `neutral', in the context of the image. We use `true', `false' and `neutral' as the output words corresponding to entailment, contradiction or neutral.

\textbf{Natural Language for Visual Reasoning (NLVR2).} NLVR2~\cite{suhr2017acorpus} dataset requires reasoning about two images simultaneously. Challenging questions include counting if the left image contains n times more objects of a certain type. We use the NLVR2 version which contains real images, 107,292 text annotations. Here we use `true' and `false' as the target words.

\textbf{Visual Genome-Question Answering (VG-QA).} The VG-QA dataset is based on Visual Genome. It contains 1.7M question-answer pairs, with %an average of 
$\sim$17 question-answers per image. We use the ground truth answers as the tokens.

\textbf{GQA.} The GQA dataset~\cite{hudson2019gqa} is also based on the Visual Genome dataset~\cite{krishnavisualgenome} and introduces more complex compositional questions based on scene object relationships. It consists of $\sim$22M question-answer pairs for 113k images.

% \textbf{Visual Commonsense Reasoning (VCR).} The VCR dataset~\cite{zellers2019vcr} consists of 110k images and provides 290k multiple choice questions and 290k answers and rationales, one rationale per answer. The dataset includes cognition-based questions, which require certain reasoning about people, objects and relationships in the image. 

%\textbf{Conceptual Captions (CC12m).} The CC12m
\textbf{CC12m.} The Conceptual Captions 12m dataset~\cite{changpinyo2021telling} is an image captioning dataset collected from the web, containing 12M image-text pairs.
This is our only pretraining dataset.

The datatsets considered in this work are collected from various sources e.g., web images, Flickr, personal mobile phones, etc.

\vspace{-0.15cm}
\subsection{Evaluation}
%\subsection{Evaluation protocols}

For each of the above-mentioned datasets above, we follow the evaluation protocols established in prior work. We also use standard well adopted metrics to measure performance. For example, accuracy on SNLI-VE, NLVR2 and GQA, the VQA2.0 accuracy metric on VQA2.0 (dev set). Notably, instead of training a classification layer for each dataset, we use a large, open vocabulary and generate text answers which are used to compute the metrics. %Please see the sup. material for details. %on the metrics.

\textbf{Zero-Shot Evaluation}: In our experiments below we conduct Zero-Shot (ZS) evaluation across tasks. We use the standard evaluation metrics per each dataset, where in the ZS setting the model is trained on different datasets. In some cases, datasets might stem from the same source e.g., GQA and VG-QA or have any overlap, in which case %we do not evaluate in ZS. We also 
we ensure training and validation/test sets do not contain overlapping images across all our experiments.

\vspace{-0.2cm}
\section{Experimental results}
\vspace{-0.05cm}
% moved from above
%After the pretraining, we evaluate the model in both zero-shot and finetuning settings on multiple datasets and tasks. 

Here we show the main results of the Answer-Me multi-task training. 
We evaluate the approach in a zero-shot (ZS) setting (Section~\ref{sec:zs_results}), and on multiple question-answering tasks. We then train and evaluate with multiple task mixtures (Section~\ref{sec:main_results}), and demonstrate Answer-Me is beneficial against catastrophic forgetting when using  multiple datasets (Section~\ref{sec:across}).
We also evaluate Answer-Me on an entirely new %, not related to question-answering,
task of detection conveyed via text (Section~\ref{sec:detection}), which does not fit any of the question-answering tasks and demonstrate Answer-Me generalization capabilities in both zero-shot and within-task settings.% Section~\ref{sec:ablations} has ablations.

\textbf{Experimental results summary.}
In our main experiments we evaluate the performance on Answer-Me when 4 visual question-answering tasks are combined and trained together, and when 8 tasks are combined in training. Across both zero-shot and standard multi-task experiments, we observe that the mixtures perform well on novel tasks and within-tasks, and that a mixture of more tasks results in better accuracies across all datasets tested. We further demonstrate the performance of our model when scaled 3x (has 3 times more parameters), which is easy to do with our model. Here again, across evaluation settings and across datasets, Answer-Me brings in large improvements in performance.
We demonstrate that the Answer-Me mixtures are much more robust to catastrophic forgetting and that the Answer-Me mixtures perform well across multiple tasks, instead of having higher accuracy on a single task only. 
We note that these results are also obtained with the open-vocabulary outputs from the Answer-Me model, unlike all prior work, which to our knowledge uses smaller and fixed vocabulary to evaluate their performance.
Furthermore, on the new task of detection conveyed via text, which is not a question answering task, we observe the utility of the approach on this unique and novel task, showing its ability to recognize a variety of objects, even in a zero-shot setting.

\begin{table}[t]
    \centering
      \caption{\textbf{Zero-shot} performance of multi-task Answer-Me for mixtures of tasks. The mixtures do not include the task that is being tested on (we make sure there is no `leakage' to the test set for each experiment). As seen, the mixture improves zero-shot performance over pretraining. Increasing the Answer-Me mixture set (number of tasks in the mixture) leads to better ZS performance. Scaling the model (last row) is additionally beneficial across all tasks.}
    \label{tab:zero-shot}
    \scalebox{0.99}{  % adjust as needed
    \hspace*{-3mm}\begin{tabular}{l|cccc}
    \toprule
     Approach &VQA  &NLVR2  &SNLI-VE  &GQA \\
    \midrule
%Single task (not zero-shot)  &   &  & & \\    
%\midrule
%Frozen \cite{frozen} & 29.5 & - & - & - \\
Pretrained only &25.3   &32.5  &22.7 &40.9 \\    
Answer-Me 4-ZS-tasks  &30.0   &42.5   &34.1  &42.3 \\  
Answer-Me 8-ZS-tasks  &\textbf{35.0} &\textbf{44.7}  &\textbf{37.3} &\textbf{44.2} \\
Answer-Me 8-ZS-tasks, 3x & \textbf{39.2} &\textbf{ 48.3} & \textbf{41.1} & \textbf{47.2} \\
    \bottomrule
    \end{tabular}
    }
\end{table}

\begin{table}[t]
    \centering
    \caption{Experiments comparing Answer-Me multi-task learning, evaluated on several datasets. As seen, more tasks improve performance across all datasets; scaling the model, which is easy in our framework, brings in further consistent improvements. Results from models, fine-tuned to individual tasks, are shown in the top portion of the table, multi-task models (a single one per evaluation), are shown in the bottom portion.
    %Last two tasks are Out-of-Distribution tasks (i.e. are not used in the mix). Note: we'll reshuffle the tasks. Using the `base' model.
    }
    \label{tab:main}
    \scalebox{0.99}{  % adjust as needed
    \hspace*{-4mm}\begin{tabular}{l|c|c|c|c|c}
    \toprule
     Approach &Num Models & VQA2.0 &NLVR2 &SNLI-VE  &GQA \\ % &VizWiz\\
    \midrule

Single-Task (random init) & Mult & 49.05 & 53.5  & 73.1 & 68.9  \\
Single-task, pretrained (PT) & Mult & 65.2 & 70.2 & 77.72 & 73.03  \\
Single-Task, PT, 3x scaled  & Mult & 71.2 &72.0 & \textbf{85.8} & 77.2  \\
%Single-Task FT from multi-task pretr  (ours - STRETCH) &Mult &  & & & & & & & &\\

\midrule 
Answer-Me, PT, Zero-shot &Single & 25.3  & 32.5 & 22.7 & 40.9  \\
Answer-Me, PT, 4 tasks &Single & 64.8  & 71.5  & 77.2 & 72.1 \\
Answer-Me, PT, 8 tasks &Single & 65.1  & 71.7  & 77.5 & 72.8 \\
Answer-Me, PT, 8 tasks, 3x &Single &\textbf{ 73.6} & \textbf{73.9} &  \textbf{85.8}  & \textbf{77.5} \\

%Zero-shot, pretrained + 4 tasks &Single &30.0   &42.5   &34.1  &42.3  \\
%Zero-shot, pretrained + 8 tasks  &Single & 35.0 & 44.7 & 37.3 & 44.2 &  \\
%Zero-shot, pretrained + 8 tasks 1B params &Single & 39.2 & 48.3 & 41.1 & 47.2 &  \\
    \bottomrule
    \end{tabular}
    }
 
    \vspace{-0.5cm}
\end{table}

\begin{table}[]
    \centering
      \caption{Comparing mixture training vs. individual task FT.}
    \label{tab:mix-vs-one}
    \scalebox{0.99}{ 
    \begin{tabular}{l|cccc}
    \toprule
        Tasks & VQA2.0 & NLVR2 & SNLI-VE & GQA \\
    \midrule
        VQA2.0-only & \textbf{65.2} & 34.2 & 24.3 & 41.3 \\
        NLVR2-only & 23.4 & \textbf{70.2} & 21.3 & 36.7 \\
        SNLI-VE-only & 24.3 & 34.6 & \textbf{77.7} & 38.5 \\
        GQA-only &  28.6 & 33.5 & 23.4 & \textbf{73.0}\\
        
        VG-QA-only (ZS) & 29.4 & 33.7 & 27.5 & 40.8 \\
        
       %  VizWiz-only & \\
    \midrule
        5-task mix & 65.1 & 71.5 & 77.4 & 72.8 \\
    \bottomrule
    \end{tabular}
    }
    \vspace{-4mm}
\end{table}

\subsection{Zero-shot results}
\label{sec:zs_results}

In \autoref{tab:zero-shot}, we evaluate Answer-Me when training on a mix of tasks in a Zero-Shot (ZS) setting, i.e., on unseen tasks. This is important as it is not always feasible to obtain full labeling of examples for new tasks. The mixture training is able to give better answers than just the pretrained model. Our results also indicate that larger mixtures lead to better zero-shot results, which shows an important ability to perform new skills.

Using the five VQA datasets above (VQA2.0, GQA, SNLI-VE and VG-QA) in the mixture setting, we mix by picking four for training and one for evaluation. The dataset evaluated on is not used in training. We further make sure any evaluation images are  excluded from the training data. 
For example, when evaluating on VQA2.0, we train on GQA, SNLI-VE and VG-QA with the VQA2.0 validation images removed from those datasets. Similarly when evaluating on SNLI-VE, we train on VQA2.0, GQA and VG-QA.
For the 8-task mixture, we additionally add Coco Captioning data~\cite{chen2015cococaptions}, Localized Narratives \cite{localizednarratves}, Visual Genome region descriptions~\cite{krishnavisualgenome}, and ImageNet matching (e.g., `Is this X?' with a `yes/no' output, generated from ImageNet classes~\cite{imagenet}). We similarly remove overlapping images from the evaluation data.% They are described in the supp.

We evaluate in zero-shot manner on VQA2.0, SNLI-VE, GQA and NLVR2. NLVR2 needs two images as input, unlike all other datasets, so here it is only used as a zero-shot evaluation. \autoref{tab:zero-shot} shows the zero-shot results, using the pretrained only, 4-task and 8-task model. The 8-task mix improves by almost 10\% on VQA2.0. 
We also find that 8 tasks mixture produces consistently better results across all datasets than 4 tasks, even though the additional 4 tasks are less related to VQA data. When scaling, we observe even higher results on VQA and consistent improvements on all datasets. 
% We also want to test out if  more tasks in the mix leads to better generalization on a held-out set.
%We also see improved zero-shot performance across tasks from a 4-mixture to an 8-mixture tasks, e.g. from 30 to 35 on VQA.
For comparison, a single model from scratch trained on VQA2.0 only achieves 49\% accuracy. The multi-task learning by leveraging other image-text datasets is able to reduce the gap to supervised training. This is very promising as it demonstrates a level of generalization to novel/unseen tasks. This further improves upon the results in Frozen \cite{frozen} which got 29.5\% for ZS on VQA.

%\subsection{Main results: Multi-task mixture results}
%\subsection{Skill transfer: Novel task experiments}
%\subsection{Main results: Novel task experiments}
\vspace{-0.2cm}
\subsection{Multi-task results}
\label{sec:main_results}
\vspace{-0.1cm}
In this section we test the capabilities of the Answer-Me models and their potential for skill transfer. I.e., we compare how a model performs when a task is included in the training mix vs. a task outside the mix.~\autoref{tab:main} compares Answer-Me trained on single tasks vs. different task mixtures. We observe that the mixtures provide competitive results to single pretrained and fine-tuned (FT) models, and that more tasks in the mixture improve the performance across all datasets. When scaling the model, we see consistent improvements. In \autoref{tab:mix-vs-one} we compare single-task vs. multi-task training, then evaluating on all 4 different VQA datasets. We see that when fine-tuning on a single task, the model does well on that task, but poorly on the others, but when trained on a mix of all 5 VQA datasets in the table, the model does nearly the same as the single task training. Together, these tables show robustness to forgetting and generalization ability to a variety of question types.

In \autoref{tab:sota} we compare to the state-of-the-art (SOTA) results. 
We compare to other multi-task state-of-the-art methods, such as UniT~\cite{hu2021unit}, 12-in-1~\cite{lu202112in1}, GPV~\cite{gupta2021gpv}, and others. 

We note that our results are obtained in the open-vocabulary setting which is more challenging, as (in our case) the generated text has to match exactly the ground truth to be counted as accurate. This is additionally compounded by the multi-task setting as the method is less encouraged to memorize answers within a task or datasets. While more challenging, this is a more realistic scenario and we hope that other works commence adopting it.

Despite the more challenging setting, our results largely outperform other SOTA approaches and for some datasets, e.g. GQA, SNLI-VE, with large margins. Only 12-in-1~\cite{lu202112in1} outperforms our approach and only on the NLVR2 dataset. This demonstrates the generalization advantages of our multi-task approach.

%Of note is that some of the multi-task methods, e.g. UniT and 12-in-1 use task-specific heads, while our model performs similarly to many previous works, while using a simpler architecture.

%We note that the SOTA results are usually individual models or task heads specific to each task or dataset. 
Furthermore, our approach compares well even with additionally fine-tuned methods, which are further advantaged by fine-tuning to a specific task. Specifically, Answer-Me outperforms all prior fine-tuned approaches on GQA datasets,
%Some are individual models, fine-tuned over strong pretraining, in some cases from strong mixture of tasks and datasets, e.g., UNITER~\cite{chen2020uniter}, Oscar~\cite{li2020oscar}, VL-Bert~\cite{vilbert2020}, etc.
and outperforms some of them on all datasets, e.g.
 VisualBERT \cite{li2019visualbert},
 ViLBERT \cite{vilbert2020}.
SimVLM-Huge~\cite{simvlm} which is a 1.5 Billion parameter model trained on the 1.1 Billion image-language Align dataset~\cite{align} largely performs best among these. Note that our multi-task open-vocabulary model is close in performance to SimVLM-Huge and outperforms SimVLM-Large on SNLI-VE, despite using smaller data and smaller model, which further demonstrates the power of multi-task diverse training for VQA. 
% On par with LXMERT (pretr+FT) \cite{tan2019lxmert} &  69.9 &74.9 &- & 60.0& 55.4 \\
% Oscar (pretr+FT) \cite{li2020oscar} &  73.82 & - & - & - & - \\
% VinVL (pretr+FT) \cite{zhang2021vinvl} &  \textbf{75.95} & \textbf{82.05} &- & 65.05 &  \\
% Uniter (pretr+FT) \cite{chen2020uniter} & 72.70 & 77.18 & 79.39 & - & -  \\
 
%

As seen in \autoref{tab:zero-shot}, \autoref{tab:main} and \autoref{tab:sota}, there is progressive improvement with adding more tasks in the mixture and scaling the model, here 3X, produces consistently higher accuracies.

\begin{table*}[t]
    \centering
      \caption{Experiments comparing to SOTA: specialized models (top section), and multi-task models, including ours (middle large section). Answer-Me largely outperforms other multi-task models despite working in the open-vocabulary generative setting. For reference we include pre-trained and fine-tuned models which are further advantaged by fine-tuning to each individual dataset (bottom section). As seen, Answer-Me still outperforms these on GQA and it even outperforms the Large version of SimVLM model~\cite{simvlm} and it is close to its SimVLM-Huge on SNLI-VE.  
      The best results among the pre-trained fine-tuned models are marked in italics.}
    \label{tab:sota}
    \scalebox{0.9}{  % adjust as needed
    \begin{tabular}{l|c|c|c|c}
    \toprule
     Approach & VQA2.0 (dev)& NLVR2 &SNLI-VE   &GQA   \\
    \midrule
DFAF~\cite{dfaf}  &70.22  &- &- &-   \\
Specialized from ~\cite{gupta2021gpv} &60.1 &- &- &-  \\
Suhr et al.~\cite{suhr2017acorpus}  &- &53.5 &- & - \\
Xie et al.~\cite{snli-ve}  &- &- &71.56 &-    \\ 
Hudson \& Manning~\cite{hudson2019gqa}  &- &- &- &57.5   \\
%\midrule

\midrule
% \midrule 

Frozen~\cite{frozen} (VQA+pretraining) &48.4 &- &- &-   \\
\midrule
  Nguyen et al~\cite{nguyen2019multi} (VQA2.0+VG) &66.35 &- &- &-   \\
  Multi-task GPV ~\cite{gupta2021gpv} &62.5 &- &- &-   \\
  VL-BART \cite{cho2021unifying} & 71.3 & 70.3 & - & 60.5   \\
  VL-T5 \cite{cho2021unifying} & 70.3 & 73.6 & - & 60.8    \\
  12-in-1~\cite{lu202112in1}  &72.57    &\textbf{78.4} &76.78  &60.12  \\
  UniT (Coco init.)~\cite{hu2021unit} &66.97   &- &73.16  
  &-   \\
\midrule 
 \textbf{Answer-Me}, 8 tasks (Ours) & 65.1 & 71.7 & 77.5 & 72.8 \\
 \textbf{Answer-Me}, 8 tasks, 3x scaling (Ours) & \textbf{73.6} & 73.9 & \textbf{85.8} & \textbf{77.5}  \\
% Answer-Me, 3x, above + FT & 73.8 & 74.2 & \textbf{86.1} & \textbf{77.7} & \textbf{75.8} \\
\midrule
\midrule
VisualBERT (pretr+FT) \cite{li2019visualbert}            &67.36  & 66.7 & 75.69 & -  \\
 ViLBERT (pretr+FT) \cite{vilbert2020} & 70.55 & - & - & -\\
 LXMERT (pretr+FT) \cite{tan2019lxmert} &  69.9 &74.9 &- & 60.0\\
% ViLT (pretr+FT) \cite{kim2021vilt} & 71.26	& 75.7 & - & - \\
 
 %SNLI-VE (pretr+FT, Align) &   &? &73.56 & &? & &  \\
Oscar (pretr+FT) \cite{li2020oscar} &  73.61 & - & - & - \\
 Uniter (pretr+FT) \cite{chen2020uniter} & 73.82 & 79.12 & 79.39 & -   \\
 VinVL (pretr+FT) \cite{zhang2021vinvl} &  75.95 & {\it 82.05} &- & {\it 65.05}   \\
 SimVLM (Large) (pretr+FT) \cite{chen2020uniter} & 79.32 & 84.13 & 85.68 & -   \\
SimVLM (Huge) (pretr+FT) \cite{chen2020uniter} & {\it 80.03} & {\it 84.53} & {\it 86.21} & -   \\

    \bottomrule
    \end{tabular}
    }
\end{table*}

\begin{table}[]
    \centering
    \caption{
    The pretrained base model trained on the 3 VQA (GQA, VG-QA) mix plus either VQA2.0 or SNLI-VE. This is evaluated on VQA2.0 and SNLI-VE. We further train the model on one of the tasks and repeat the evaluations. The results show that fine tuned models tend to forget (first/second rows), even if original mix shows good within-data and out-of sample generalization (first rows). Additional fine-tuning seems to recover the losses within a task (first/third rows), but costs $N$ times the cost in training, and performance on the other task deteriorates again. Interestingly, this model performs even worse than the original out-of-sample mixture on the second task.
    Training on many tasks in the mix maintains performance (last row).
    }
    \label{tab:oos}
    \scalebox{0.99}{  % adjust as needed
    \begin{tabular}{l|c|c||c}
    \toprule
     Approach &Num  &VQA2.0  &SNLI-VE\\
    \midrule

%\midrule
3-task + VQA2.0  (ours) &Single & 64.3  & 33.8 \\
3-task + VQAv, FT on SNLI &Multiple & 35.5  & 76.9 \\
3-task + VQA2.0, FT on VQA2.0 (ours) &Multiple  & 65.2  & 26.7 \\
\midrule
3-task + SNLI-VE  (ours) & Single & 33.2  & 76.5 \\
3-task + SNLI-VE, FT on VQA2.0 & Multiple  & 64.8  & 24.5 \\
3-task + SNLI-VE, FT on SNLI (ours) & Multiple & 29.4  & 77.2 \\
%\midrule
%3-task + VQA  (ours) &Single & 64.3  & 33.8 \\
%3-task + VQA, FT on SNLI &Multiple & 35.5  & 76.9 \\
%3-task + VQA, FT on VQA (ours) &Multiple  & 65.2  & 26.7 \\

\midrule
Multi-task (w/o VQA2.0 or SN) Zero-Shot (ours) &Single  & 27.3  & 24.2 \\
5-task (ours) & Single  & 65.1  & 77.4 \\
%\midrule
%Eval(Zero-shot, no FT) from pretraining &Single & & & & & & & & \\
%Eval(no FT) from multi-task pretr (ours) &Single & & & & & & & & \\
%Eval(no FT) from multi-task pretr+pretr (ours) &Single &  & & & & & & & \\

%\midrule 
    \bottomrule
    \end{tabular}
    }
    
\end{table}

% VQA is dev set

%TODO: how to handle uniter, etc?
%

\vspace{-0.2cm}
%\subsection{Learning across tasks}
\subsection{Answer-Me prevents catastrophic forgetting}
\label{sec:across}

While pretraining and fine-tuning, as is customarily done in previous works, produces accurate models, it tends to overfit to the new data and to immediately forget other datasets or tasks, even when it was previously trained on them. We show that Answer-Me, through the mixture training, is more robust, as it is able to sustain good performance across tasks. In \autoref{tab:oos}, we find that when training on one task, then finetuning on a second (as commonly done in previous works), the model does well on the new task, but performs poorly on the first as it `forgets', achieving accuracies close to the zero-shot model; additional fine-tuning on the first task actually makes the performance on the second one even worse than zero-shot. In contrast, when using the mixture training for Answer-Me, the model maintains the single-task performance for both datasets.

\vspace{-0.2cm}
\subsection{Novel semantic `detection' task}
\label{sec:detection}
We also examine Answer-Me's ability to `detect' objects through language. This is quite different from the other tasks presented in the paper, and it shows the model's ability to understand the whole image and many objects present in it, even in a zero-shot setting. The task is to output as text the names of all the objects in the image. If an object appears multiple times, e.g., 3 times, then the model should output the name 3 times. We evaluate this on the standard MsCoco detection dataset \cite{mscoco} and compute the precision, recall and $F_1$ values based on how many output object names match the ground truth objects. This is challenging as it requires localization awareness of the models, but without extra box regression layers and training that are not part of Answer-Me. It is also unique with respect to other question-answering tasks. We here evaluate this task in a zero-shot and full mixture scenarios. \autoref{tab:detection} shows that the Answer-Me model and training is able to do well both in zero-shot and FT settings.

\begin{table}[]
    \centering
    \caption{`Detection' task, conveyed by text, where the desired outcome is to list (in text) all objects present in the image. This evaluation is on the MsCoco detection dataset~\cite{mscoco}. As seen, Answer-Me mixture of question-answering tasks helps boost this new detection task beyond MsCoco. %, and larger mixture performs better. 
    The Zero-Shot portion does not use MsCoco data for training. %The X- and Y-task mix are two different mixtures of tasks.
    The (A) and (B) task mix are different 4-task mixtures; they both have VQA2.0, SNLI-VE, GQA, 4-task(A) has VG-QA, whereas 4-task(B) has VG region descriptions instead, which is clearly much more advantageous.
    % The X-task mix has VQA2.0, SNLI-VE, GQA and VG-QA and Y-task mix has VQA2.0, VG region description, SNLI-VE, and GQA.
    }
    \label{tab:detection}
    \scalebox{0.99}{
    \begin{tabular}{l|ccc}
    \toprule
        Train Data & Recall & Precision & F1 \\
    \midrule
        MsCoco (from scratch) & 0.25 & 0.21 & 0.22 \\
        MsCoco & 0.31 & 0.24 & 0.27 \\
        4-task (A) mix + MsCoco & 0.34 & 0.30 & 0.31 \\
        4-task (B) mix + MsCoco & \textbf{0.55} & \textbf{0.52} & \textbf{0.53} \\
    \midrule
    \multicolumn{3}{l}{Zero-shot} \\
    \midrule
      pretraining  & 0.03 & 0.02 & 0.02 \\
      4-task (A) mix & 0.06 & 0.04 & 0.05 \\
      4-task (B) mix &\textbf{ 0.20} & \textbf{0.14} & \textbf{0.16} \\
    \bottomrule
    \end{tabular}
    }
  
    \vspace{-0.2cm}
\end{table}

\begin{figure*}[t]
    \centering
    \includegraphics[width=\linewidth]{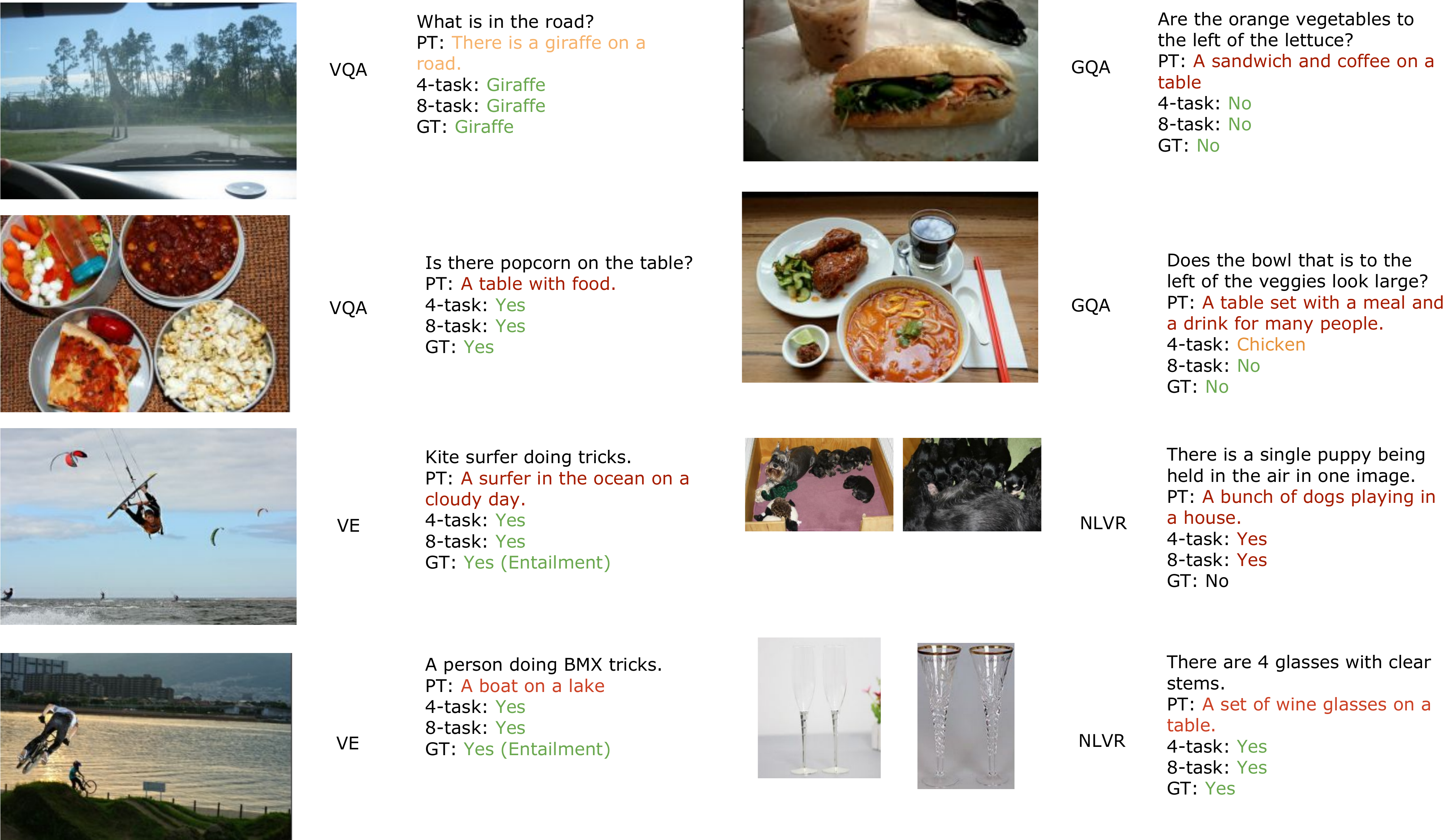}
    \caption{Example Answer-Me results, both successful and unsuccessful. We note that training for question-answering tends to tweak otherwise well sounding image captions to answer the questions more adequately.
    }
    
    \label{fig:results}
\end{figure*}

\vspace{-0.2cm}
\subsection{Ablations}
\label{sec:ablations}

%\textbf{pretraining}.
\autoref{tab:pretraining-tasks} shows the performance of various tasks for pretraining. We find that using all 4 tasks provides the best model, as it is able to better train all parts of the model, including the decoder which is often ignored in contrastive-style pretraining and the encoder ignored in captioning pretraining. Adding these tasks is simple as they all use the same loss, just slightly different preprocessing.

%\subsection{Human judgement}
%Here we evaluate our proposed approach %(maybe on a %held-out set, or any) 
%vs the baseline using human ratings, as descibed above.
%TODO: Results? + Table

\begin{table}[]
    \centering
        \caption{Study on different tasks for pretraining (individually fine-tuned). Using all four pretraining tasks is best, and outperforms any one of them used alone, in some cases by large margins.}
    \label{tab:pretraining-tasks}
    \scalebox{0.99}{
    \begin{tabular}{l|cc}
    \toprule
    PT Task & VQA2.0 & SNLI-VE  \\
    \midrule
     Captioning        & 62.3 & 75.2   \\
     CapCompletion     & 60.1 & 73.8  \\
     ITM               & 54.5 & 74.2 \\
     MLM               & 58.3 & 72.4 \\
     All 4             & \textbf{65.2} & \textbf{77.7} \\
     \bottomrule
    \end{tabular}
    }
    \vspace{-6mm}
\end{table}

% one stream vs two stream ablation
We also experiment with an alternative architectural choice for fusion in which a transformer encoder/decoder layer is used instead of concatenation and transformer encoders. In this alternative approach, the queries come from the text input and the key/value comes from the image input. 
%, and both self-attention and encoder/decoder attention are non-causal. 
\autoref{tab:fusion-types} shows that using the encoder/decoder fusion style has little to no effect on the benchmark metrics but the number of parameters is higher compared to the encoder-only fusion. We therefore opted to use concatenation and transformer encoders for fusion. % as previously highlighted.
%Please see the supp. for more results and implementation details.

\begin{table}[]
\centering
    \caption{Study on different fusion layer types. Both approaches behave similarly but encoder fusion has fewer parameters. The experiment is done with no pretraining, so the numbers are lower. }
    \label{tab:fusion-types}
    \scalebox{0.99}{
    \begin{tabular}{l|ccc}
    \toprule
    Fusion Approach & VQA2.0 & SNLI-VE & Num Params  \\
    \midrule
     Encoder             & 49.05 & 73.1 & 332M   \\
     Encoder/Decoder     & 48.9 & 73.3 & 346M \\
     \bottomrule
    \end{tabular}
    }
  
    \vspace{-4mm}
\end{table}

\textbf{Limitations.} %\hspace{0.2cm}
One limitation of our approach is that generated text outputs might not always be best, %form proper or grammatically correct outputs, 
this is likely because we use data and models of modest sizes. 
It is also unable to read text, %e.g., perform OCR,
which can be remedied from the inclusion of TextVQA datasets~\cite{singh2019towards} or large diverse datasets~\cite{clip}.  Other limitations are related to measuring the performance of these models. %In our best efforts we included human-rated performance evaluation, but 
The evaluation can be further refined and extended, by including human judgement. %TODO 
%We acknowledge that the evaluation can be further refined and extended, e.g., to rate potentially offensive, inadequate or biased responses. %TODO check if human exp makes it.

% Added elsewhere
%Complexity of questions

%Complexity of scenes

% Data bias and fairness, added below

\textbf{Societal impact.} 
Multi-modal image-language models in conjunction with large datasets can potentially use training examples with biased or offensive text or images. While best efforts have been applied in the collection of the Conceptual Captions dataset~\cite{beer2021cc12m}, which we use as a pretraining dataset, we are aware of the potential pitfalls of large image-language corpora with regards to bias, fairness and safety.
Our mitigation strategy is to not share, release or otherwise deploy models trained on this data 
and use these results for evaluating the capabilities of the proposed methods and technologies, to facilitate future research in this direction. 

%\section{Conclusion and future work}
\section{Conclusions}
We introduce Answer-Me, a framework for multi-task training and pretraining, for answering natural language questions towards an image with natural responses for multiple tasks, such as VQA, visual reasoning, and others. It can answer questions from multiple diverse tasks seamlessly, understanding the intent of the question without specific task specification or additional prompts, and can generalize to novel tasks.

We show Answer-Me is capable of zero-shot QA tasks and is robust to forgetting on a variety of tasks. The model is simple, has an easily scalable and extensible architecture which learns the image-language interactions across many tasks.

Answer-Me performs well with respect to SOTA, largely outperforming prior multi-task models, and in some cases the fine-tuned (and sometimes larger) models. It is being evaluated in the open-vocabulary setting which is a more challenging one, demonstrating the power of our approach for more practical scenarios. We hope that other methods adopt the open-vocabulary generative setting in evaluation in future work, as well.

\clearpage
\bibliographystyle{splncs04}
\bibliography{egbib}

\begin{thebibliography}{10}
\providecommand{\url}[1]{\texttt{#1}}
\providecommand{\urlprefix}{URL }
\providecommand{\doi}[1]{https://doi.org/#1}

\bibitem{agrawal2015vqa}
Agrawal, A., Lu, J., Antol, S., Mitchell, M., Zitnick, C.L., Batra, D., Parikh,
  D.: Vqa: Visual question answering. In: ICCV (2015)

\bibitem{anderson2018bottomup}
Anderson, P., He, X., Buehler, C., Teney, D., Johnson, M., Gould, S., Zhang,
  L.: Bottom-up and top-down attention for image captioning and visual question
  answering. In: CVPR (2018)

\bibitem{snli}
Bowman, S.R., Angeli, G., Potts, C., Manning, C.D.: A large annotated corpus
  for learning natural language inference. In: EMNLP (2015)

\bibitem{detr}
Carion, N., Massa, F., Synnaeve, G., Usunier, N., Kirillov, A., Zagoruyko, S.:
  End-to-end object detection with transformers. In:
  https://arxiv.org/abs/2005.12872 (2020)

\bibitem{beer2019decoupled}
Changpinyo, S., Pang, B., Sharma, P., Soricut, R.: Decoupled box proposal and
  featurization with ultrafine-grained semantic labels improve image captioning
  and visual question answering. In: Conference on Empirical Methods in Natural
  Language Processing (EMNLP) (2019)

\bibitem{changpinyo2021telling}
Changpinyo, S., Pont-Tuset, J., Ferrari, V., Soricut, R.: Telling the what
  while pointing to the where: Multimodal queries for image retrieval. In:
  Arxiv 2102.04980 (2021)

\bibitem{beer2021cc12m}
Changpinyo, S., Sharma, P., Ding, N., Soricut, R.: Conceptual 12m: Pushing
  web-scale image-text pre-training to recognize long-tail visual concepts. In:
  CVPR (2021)

\bibitem{chen2015cococaptions}
Chen, X., Fang, H., Lin, T.Y., Vedantam, R., Gupta, S., Dollar, P., Zitnick,
  C.L.: Microsoft coco captions: Data collection and evaluation server. In:
  https://arxiv.org/abs/1504.00325 (2015)

\bibitem{chen2020uniter}
Chen, Y.C., Li, L., Yu, L., Kholy, A.E., Ahmed, F., Gan, Z., Cheng, Y., Liu,
  J.: Uniter: Universal image-text representation learning. In: ECCV (2020)

\bibitem{cho2021unifying}
Cho, J., Lei, J., Tan, H., Bansal, M.: Unifying vision-and-language tasks via
  text generation. In: Arxiv 2102.02779 (2021)

\bibitem{das2017Visual}
Das, A., Kottur, S., Gupta, K., Singh, A., Yadav, D., Moura, J.M.F., Parikh,
  D., Batra, D.: Visual dialog. In: CVPR (2017)

\bibitem{deng2021sketch}
Deng, C., Chen, S., Chen, D., He, Y., Wu, Q.: Sketch, ground, and refine:
  Top-down dense video captioning. In: CVPR (2021)

\bibitem{imagenet}
Deng, J., Dong, W., Socher, R., Li, L.J., Li, K., Fei-Fei, L.: Imagenet: A
  large-scale hierarchical image database. In: CVPR (2009)

\bibitem{deng2021transvg}
Deng, J., Yang, Z., Chen, T., Zhou, W., Li, H.: Transvg: End-to-end visual
  grounding with transformers. ICCV  (2021)

\bibitem{virtex}
Desai, K., Johnson, J.: Virtex: Learning visual representations from textual
  annotations. In: CVPR (2021)

\bibitem{devlin2018bert}
Devlin, J., Chang, M.W., Lee, K., Toutanova, K.: Bert: Pre-training of deep
  bidirectional transformers for language understanding. arXiv preprint
  arXiv:1810.04805  (2018)

\bibitem{villa}
Gan, Z., Chen, Y.C., Li, L., Zhu, C., Cheng, Y., Liu, J.: Large-scale
  adversarial training for vision-and-language representation learning. In:
  NeurIPS (2020)

\bibitem{goyal2017making}
Goyal, Y., Khot, T., Summers-Stay, D., Batra, D., Parikh, D.: Making the {V} in
  {V}{Q}{A} matter: Elevating the role of image understanding in visual
  question answering. In: CVPR (2017)

\bibitem{gupta2021gpv}
Gupta, T., Kamath, A., Kembhavi, A., Hoiem, D.: Towards general purpose vision
  systems. In: arxiv.org/abs/2104.00743 (2021)

\bibitem{gurari2018vizwiz}
Gurari, D., Li, Q., Stangl, A.J., Guo, A., Luo, C.L.K.G.J., Bigham, J.P.:
  Viz{W}iz grand challenge: Answering visual questions from blind people. In:
  CVPR (2018)

\bibitem{hu2021unit}
Hu, R., Singh, A.: Unit: Multimodal multitask learning with a unified
  transformer. In: arxiv.org/abs/2102.10772 (2021)

\bibitem{huang2020multimodal}
Huang, G., ~, B.P., Zhu, Z., Rivera, C., Soricut, R.: Multimodal pretraining
  for dense video captioning. In: AACL-IJCNLP (2020)

\bibitem{huang2021seeing}
Huang, Z., Zeng, Z., Huang, Y., Liu, B., Fu, D., Fu, J.: Seeing out of the
  box:end-to-end pre-training for vision-language representation learning. In:
  CVPR (2021)

\bibitem{hudson2019gqa}
Hudson, D.A., Manning, C.D.: Gqa: a new dataset for compositional question
  answering over realworld images. In: CVPR (2019)

\bibitem{align}
Jia, C., Yang, Y., Xia, Y., Chen, Y.T., Parekh, Z., Pham, H., Le, Q.V., Sung,
  Y., Li, Z., Duerig, T.: Scaling up visual and vision-language representation
  learning with noisy text supervision. In: ICML (2021)

\bibitem{jiang2020in}
Jiang, H., Misra, I., Rohrbach, M., Learned-Miller, E., Chen, X.: In defense of
  grid features for visual question answering. In: CVPR (2020)

\bibitem{mdetr}
Kamath, A., Singh, M., LeCun, Y., Misra, I., Synnaeve, G., Carion, N.: Mdetr -
  modulated detection for end-to-end multi-modal understanding. In:
  https://arxiv.org/abs/2104.12763 (2021)

\bibitem{kim2021vilt}
Kim, W., Son, B., Kim, I.: Vilt: Vision-and-language transformer without
  convolution or region supervision. In: ICML (2021)

\bibitem{kottur2018Visual}
Kottur, S., Moura, J.M.F., Parikh, D., Batra, D., Rohrbach, M.: Visual
  coreference resolution in visual dialog using neural module networks. In:
  ECCV (2018)

\bibitem{krishnavisualgenome}
Krishna, R., Zhu, Y., Groth, O., Johnson, J., Hata, K., Kravitz, J., Chen, S.,
  Kalantidis, Y., Li, L.J., Shamma, D.A., Bernstein, M., Fei-Fei, L.: Visual
  genome: Connecting language and vision using crowdsourced dense image
  annotations (2016), \url{https://arxiv.org/abs/1602.07332}

\bibitem{li2019visualbert}
Li, L.H., Yatskar, M., Yin, D., Hsieh, C.J., Chang, K.W.: Visualbert: A simple
  and performant baseline for vision and language. arXiv preprint
  arXiv:1908.03557  (2019)

\bibitem{li2020oscar}
Li, X., Yin, X., Li, C., Zhang, P., Hu, X., Zhang, L., Wang, L., Hu, H., Dong,
  L., Wei, F., Choi, Y., Gao, J.: Oscar: Object-semantics aligned pre-training
  for vision-language tasks. In: ECCV (2020)

\bibitem{liao2020real}
Liao, Y., Liu, S., Li, G., Wang, F., Chen, Y., Qian, C., Li, B.: A real-time
  cross-modality correlation filtering method for referring expression
  comprehension. In: CVPR. pp. 10880--10889 (2020)

\bibitem{mscoco}
Lin, T.Y., Maire, M., Belongie, S., Hays, J., Perona, P., Ramanan, D., Dollar,
  P., Zitnick, C.L.: Microsoft {C}{O}{C}{O}: Common objects in context. In:
  ECCV (2014)

\bibitem{lin2021vx2text}
Lin, X., Bertasius, G., Wang, J., Chang, S.F., Parikh, D.: Vx2text: End-to-end
  learning of video-based text generation from multimodal inputs. In: CVPR
  (2021)

\bibitem{liu2019learning}
Liu, D., Zhang, H., Wu, F., Zha, Z.J.: Learning to assemble neural module tree
  networks for visual grounding. In: ICCV (2019)

\bibitem{liu2019multitask}
Liu, X., He, P., Chen, W., Gao, J.: Multi-task deep neural networks for natural
  language understanding. In: In Proceedings of ACL (2019)

\bibitem{liu2021enhancing}
Liu, Y., Huang, L., Song, L., Wang, B., Zhang, Y., Pan, P.: Enhancing textual
  cues in multi-modal transformers for {V}{Q}{A}. In: VizWiz Challenge 2021
  (2021)

\bibitem{loctex}
Liu, Z., Stent, S., Li, J., Gideon, J., Han, S.: Loctex: Learning
  data-efficient visual representations from localized textual supervision. In:
  ICCV (2021)

\bibitem{vilbert2020}
Lu, J., Batra, D., Parikh, D., Lee, S.: Vilbert: Pretraining task-agnostic
  visiolinguistic representations for vision-and-language tasks. In: CVPR
  (2019)

\bibitem{lu202112in1}
Lu, J., Goswami, V., Rohrbach, M., Parikh, D., Lee, S.: 12-in-1: Multi-task
  vision and language representation learning. In: CVPR (2020)

\bibitem{mao2016generation}
Mao, J., Huang, J., Toshev, A., Camburu, O., Yuille, A., Murphy, K.: Generation
  and comprehension of unambiguous object descriptions. In: CVPR (2016)

\bibitem{Tuay1018dynamic}
Margffoy-Tuay, E., Perez, J.C., Botero, E., Arbelaez, P.: Dynamic multimodal
  instance segmentation guided by natural language queries. In: ECCV (2018)

\bibitem{nguyen2019multi}
Nguyen, D.K., Okatani, T.: Multi-task learning of hierarchical vision-language
  representation. In: CVPR (2019)

\bibitem{dfaf}
Peng, G., Jiang, Z., You, H., Lu, P., Hoi, S., Wang, X., Li, H.: Dynamic fusion
  with intra- and inter- modality attention flow for visual question answering.
  In: CVPR (2019)

\bibitem{plummer2020revisiting}
Plummer, B.A., Shih, K.J., Li, Y., Xu, K., Lazebnik, S., Sclaroff, S., Saenko,
  K.: Revisiting image-language networks for open-ended phrase detection. In:
  TPAMI (2020)

\bibitem{plummer2017FlickrEntities}
Plummer, B.A., Wang, L., Cervantes, C.M., Caicedo, J.C., Hockenmaier, J.,
  Lazebnik, S.: Flickr30k entities: Collecting region-to-phrase correspondences
  for richer image-to-sentence models. In: International Journal of Computer
  Vision (2017)

\bibitem{localizednarratves}
Pont-Tuset, J., Uijlings, J., Changpinyo, S., Soricut, R., Ferrari, V.:
  Connecting vision and language with localized narratives. In: ECCV (2020)

\bibitem{pramanik2019omninet}
Pramanik, S., Agrawal, P., Hussain, A.: Omninet: A unified architecture for
  multi-modal multi-task learning. In: arxiv.org/abs/1907.07804 (2019)

\bibitem{qiao2020referring}
Qiao, Y., Deng, C., Wu, Q.: Referring expression comprehension: A survey of
  methods and datasets. In: IEEE Transactions on Multimedia (2020)

\bibitem{clip}
Radford, A., Kim, J.W., Hallacy, C., Ramesh, A., Goh, G., Agarwal, S., Sastry,
  G., Askell, A., Mishkin, P., Clark, J., Krueger, G., Sutskever, I.: Learning
  transferable visual models from natural language supervision. In: ICML (2021)

\bibitem{T5}
Raffel, C., Shazeer, N., Roberts, A., Lee, K., Narang, S., Matena, M., Zhou,
  Y., Li, W., Liu, P.J.: Exploring the limits of transfer learning with a
  unified text-to-text transformer. In: Journal of Machine Learning Research
  (2020)

\bibitem{faster}
Ren, S., He, K., Girshick, R., Sun, J.: Faster r-cnn: Towards real-time object
  detection with region proposal networks. In: Advances in Neural Information
  Processing Systems (2015)

\bibitem{Rohrbach2016grounding}
Rohrbach, A., Rohrbach, M., Hu, R., Darrell, T., Schiele, B.: Grounding of
  textual phrases in images by reconstruction. In: ECCV (2016)

\bibitem{rohrbach2015adataset}
Rohrbach, A., Rohrbach, M., Tandon, N., Schiele, B.: A dataset for movie
  description. In: CVPR (2015)

\bibitem{sariyildiz2020learning}
Sariyildiz, M.B., Perez, J., Larlus, D.: Learning visual representations with
  caption annotations. In: ECCV (2020)

\bibitem{cc3m}
Sharma, P., Ding, N., Goodman, S., Soricut, R.: Conceptual captions: A cleaned,
  hypernymed, image alt-text dataset for automatic image captioning. In: ACL
  (2018)

\bibitem{shuster2020dodecathlon}
Shuster, K., Ju, D., Roller, S., Dinan, E., Boureau, Y.L., Weston, J.: The
  dialogue dodecathlon: Open-domain knowledge and image grounded conversational
  agents. In: ACL

\bibitem{singh2019towards}
Singh, A., Natarjan, V., Shah, M., Jiang, Y., Chen, X., Parikh, D., Rohrbach,
  M.: Towards vqa models that can read. In: CVPR (2019)

\bibitem{srinivasan2021wit}
Srinivasan, K., Raman, K., Chen, J., Bendersky, M., Najork, M.: Wit:
  Wikipedia-based image text dataset for multimodal multilingual machine
  learning. In: arXiv:2103.01913 (2021)

\bibitem{suhr2017acorpus}
Suhr, A., Lewis, M., Yeh, J., Artzi, Y.: A corpus of natural language for
  visual reasoning. In: Proceedings of the 55th Annual Meeting of the
  Association for Computational Linguistics (2017)

\bibitem{sun2019videobert}
Sun, C., Myers, A., Vondrick, C., Murphy, K., Schmid, C.: Videobert: A joint
  model for video and language representation learning. In: ICCV (2019)

\bibitem{tan2019lxmert}
Tan, H., Bansal, M.: Lxmert: Learning cross-modality encoder representations
  from transformers. In: EMNLP (2019)

\bibitem{frozen}
Tsimpoukelli, M., Menick, J., Cabi, S., Eslami, S.M.A., Vinyals, O., Hill, F.:
  Multimodal few-shot learning with frozen language models  (2021),
  \url{https://arxiv.org/abs/2106.13884}

\bibitem{vaswani2017attention}
Vaswani, A., Shazeer, N., Parmar, N., Uszkoreit, J., Jones, L., Gomez, A.N.,
  Kaiser, L., Polosukhin, I.: Attention is all you need. In: NeurIPS (2017)

\bibitem{simvlm}
Wang, Z., Yu, J., Yu, A.W., Dai, Z., Tsvetkov, Y., Cao, Y.: Simvlm: Simple
  visual language model pretraining with weak supervision. In:
  https://arxiv.org/pdf/2108.10904.pdf (2021)

\bibitem{su2020vlbert}
Weijie~Su, Xizhou~Zhu, Y.C.B.L.L.L.F.W.J.D.: Vl-bert: Pre-training of generic
  visual-linguistic representations. In: ICLR (2020)

\bibitem{whitehead2021separating}
Whitehead, S., Wu, H., Ji, H., Feris, R., Saenko, K.: Separating skills and
  concepts for novel visual question answering. In: CVPR (2021)

\bibitem{snli-ve}
Xie, N., Lai, F., Doran, D., Kadav, A.: Visual entailment: A novel task for
  fine-grained image understanding. In: https://arxiv.org/abs/1901.06706 (2019)

\bibitem{xu2016mstvtt}
Xu, J., Mei, T., Yao, T., Rui, Y.: Msr-vtt: A large video description dataset
  for bridging video and language. In: CVPR (2016)

\bibitem{yang2020improving}
Yang, Z., Chen, T., Wang, L., Luo, J.: Improving one-stage visual grounding by
  recursive sub-query construction. In: ECCV (2020)

\bibitem{yang2019fast}
Yang, Z., Gong, B., Wang, L., Huang, W., Yu, D., Luo, J.: A fast and accurate
  one-stage approach to visual grounding. In: ICCV (2019)

\bibitem{gd-vcr}
Yin, D., Li, L.H., Hu, Z., Peng, N., Chang, K.W.: Broaden the vision:
  Geo-diverse visual commonsense reasoning. In: EMNLP (2021)

\bibitem{ernievil}
Yu, F., Tang, J., Yin, W., Sun, Y., Tian, H., Wu, H., Wang, H.: Ernie-vil:
  Knowledge enhanced vision-language representations through scene graph. In:
  AAAI (2021)

\bibitem{yu2018mattnet}
Yu, L., Lin, Z., Shen, X., Yang, J., Lu, X., Bansal, M., Berg, T.L.: Mattnet:
  Modular attention network for referring expression comprehension. In: CVPR
  (2021)

\bibitem{yu2016modeling}
Yu, L., Poirson, P., Yang, S., Berg, A.C., Berg, T.L.: Modeling context in
  referring expressions. In: ECCV (2016)

\bibitem{zellers2019vcr}
Zellers, R., Bisk, Y., Farhadi, A., Choi, Y.: From recognition to cognition:
  Visual commonsense reasoning. In: CVPR (June 2019)

\bibitem{zhang2021vinvl}
Zhang, P., Li, X., Hu, X., Yang, J., Zhang, L., Wang, L., Choi, Y., Gao, J.:
  Vinvl: Revisiting visual representations in vision-language models. In:
  Proceedings of the IEEE/CVF Conference on Computer Vision and Pattern
  Recognition. pp. 5579--5588 (2021)

\bibitem{zhou2020unifiedVLP}
Zhou, L., Palangi, H., Zhang, L., Hu, H., Corso, J.J., Gao, J.: Unified
  vision-language pre-training for image captioning and vqa. In: AAAI (2020)

\end{thebibliography}
\end{document}